\documentclass[conference]{IEEEtran}
\IEEEoverridecommandlockouts
\usepackage{cite}
\usepackage{amsmath,amssymb,amsfonts}
\usepackage{graphicx}
\usepackage{textcomp}
\usepackage{xcolor}
\usepackage{algorithm}

\usepackage{booktabs}
\usepackage{multirow}

\usepackage{algpseudocode}
\usepackage{bm}
\usepackage{threeparttable}
\usepackage{multicol}
\usepackage{multirow, bm}
\usepackage{makecell}

\usepackage{algpseudocode}

\def\BibTeX{{\rm B\kern-.05em{\sc i\kern-.025em b}\kern-.08em
    T\kern-.1667em\lower.7ex\hbox{E}\kern-.125emX}}
\begin{document}

\title{Wavelet Fourier Diffuser: Frequency-Aware Diffusion Model for Reinforcement Learning}

\author{\IEEEauthorblockN{Yifu Luo}
\IEEEauthorblockA{\textit{Tsinghua University}\\
China\\}
\and
\IEEEauthorblockN{Yongzhe Chang}
\IEEEauthorblockA{\textit{Tsinghua University}\\
China\\}
\and
\IEEEauthorblockN{Xueqian Wang}
\IEEEauthorblockA{\textit{Tsinghua University}\\
China\\
Corresponding Author}
}
\maketitle

\begin{abstract}
Diffusion probability models have shown significant promise in offline reinforcement learning by directly modeling trajectory sequences. However, existing approaches primarily focus on time-domain features while overlooking frequency-domain features, leading to frequency shift and degraded performance according to our observation. In this paper, we investigate the RL problem from a new perspective of the frequency domain. We first observe that time-domain-only approaches inadvertently introduce shifts in the low-frequency components of the frequency domain, which results in trajectory instability and degraded performance. To address this issue, we propose Wavelet Fourier Diffuser (WFDiffuser), a novel diffusion-based RL framework that integrates Discrete Wavelet Transform to decompose trajectories into low- and high-frequency components. To further enhance diffusion modeling for each component, WFDiffuser employs Short-Time Fourier Transform and cross attention mechanisms to extract frequency-domain features and facilitate cross-frequency interaction. Extensive experiment results on the D4RL benchmark demonstrate that WFDiffuser effectively mitigates frequency shift, leading to smoother, more stable trajectories and improved decision-making performance over existing methods.
\end{abstract}


\section{Introduction}\label{intro}
Offline reinforcement learning (RL)\cite{levine2020offline}\cite{prudencio2023survey}, where agents learn a policy from pre-collected training datasets to maximize return without direct interaction with the environment, has garnered significant attention. This paradigm is particularly suited for scenarios where real-time data collection is impractical, time-consuming or dangerous\cite{levine2020offline}. Representative applications include healthcare\cite{fatemi2022semi}, dialog system\cite{jaques2020human}, gaming\cite{schrittwieser2020mastering}, autonomous driving\cite{shi2021offline}, and embodied AI\cite{brohan2023rt}.

Traditional offline RL approaches\cite{kumar2020conservative}\cite{wu2019behavior}\cite{kostrikov2021offline}\cite{kostrikov2021offline1}\cite{ghosh2022offline}\cite{dadashi2021offline} rely on estimating the value function, which represents the discounted sum of rewards from a given state. However, these methods often suffer from instabilities due to function approximation, off-policy learning, and bootstrapping, collectively known as the ‘deadly triad’\cite{sutton2018reinforcement}. Recently, advances in generative models have inspired a new line of research that formulates RL problems as sequence modeling tasks\cite{chen2021decision}\cite{janner2021offline}\cite{janner2022planning}.

\begin{figure}[htbp]
\centerline{\includegraphics[scale=0.35]{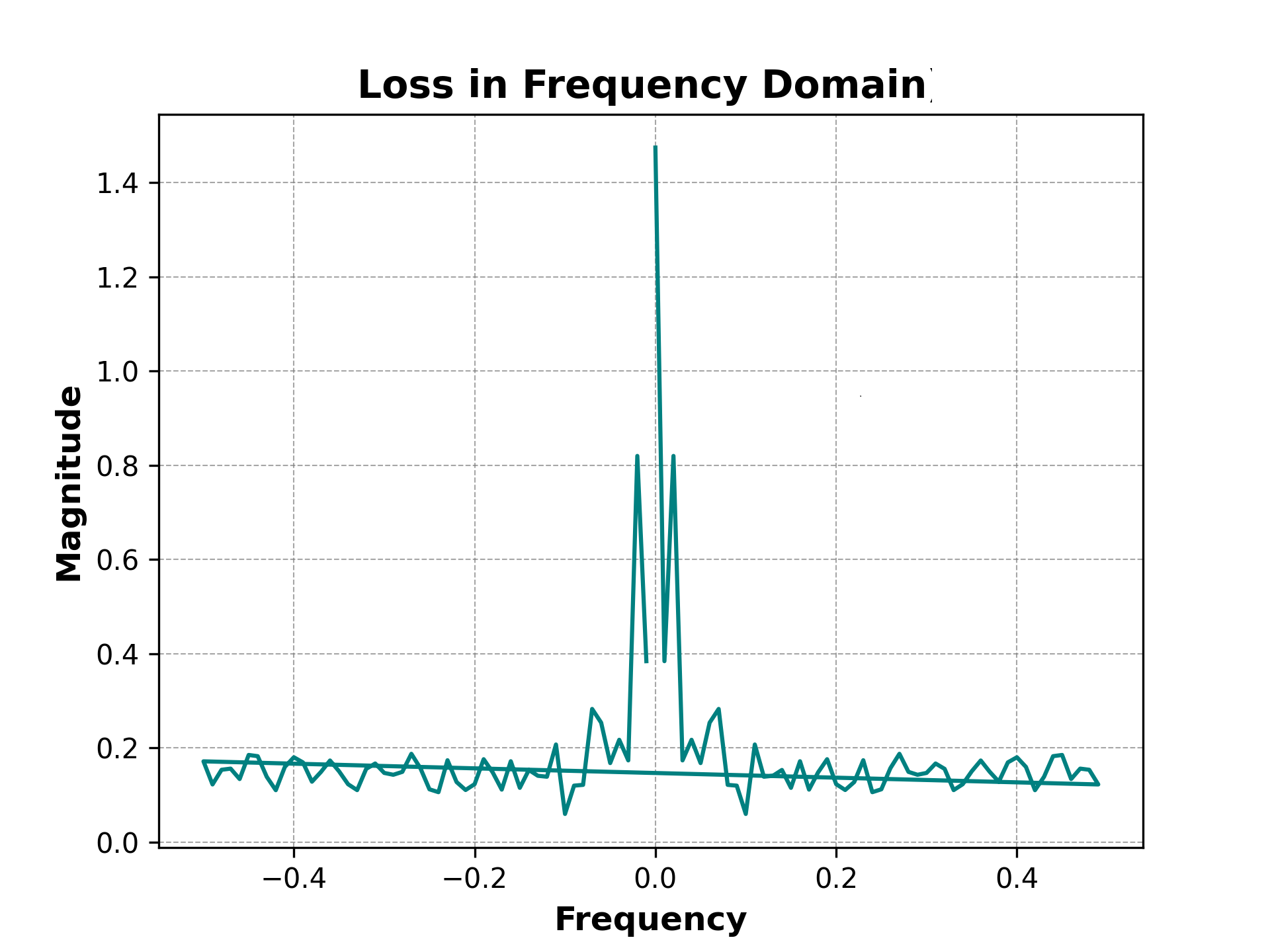}}
\caption{Using the Discrete Fourier Transform (DFT), we convert the training loss of Decision Diffuser\cite{ajay2022conditional} into the frequency domain (frequency is normalized to -0.5 $\sim$ 0.5). It is observed that the training loss is primarily concentrated in the low-frequency components (center of X-axis). The original time domain data are the joint angle training loss of Hopper-medium-v2 in D4RL dataset\cite{fu2020d4rl}}
\label{fig1}
\end{figure}

\begin{figure}[htbp]
\centerline{\includegraphics[scale=0.35]{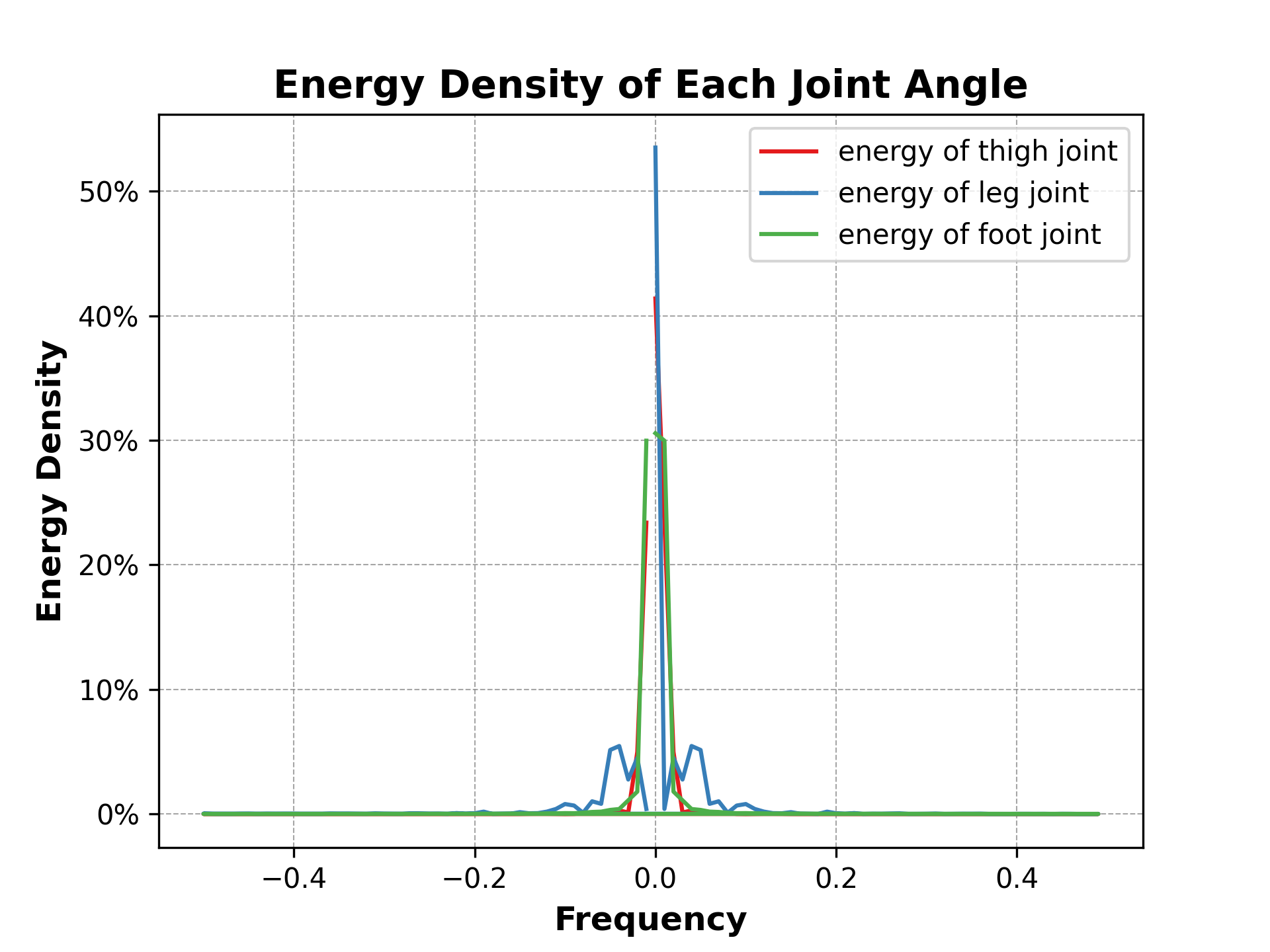}}
\caption{Energy density (normalized to 0$\%$ $\sim$ 100$\%$) in the frequency domain of a trajectory state sequence. It is observed that the energy density distribution is predominantly concentrated in the low-frequency components (center of X-axis), which shows the importance of low-frequency components to a stable trajectory. The original time domain data are the joint angle trajectory of Hopper-expert-v2 in D4RL dataset\cite{fu2020d4rl}}
\label{epoch}
\end{figure}

These approaches model the joint distribution of trajectory sequences of states, actions, and rewards to circumvent the challenge posed by the ‘deadly triad’. For instance, Decision Diffuser\cite{ajay2022conditional}, a notable example, leverages the diffusion probability models\cite{ramesh2022hierarchical}\cite{rombach2022high} to learn the distribution of trajectory sequences. During decision-making, it predicts future trajectories and employs inverse dynamics to extract and execute actions.      

Despite their promise, existing sequence modeling-based RL approaches share a common feature: they focus solely on the temporal features of trajectories in the time domain. This choice stems from an analogy to natural language modeling\cite{vaswani2017attention}\cite{austin2021structured}, where generative models have achieved remarkable success. However, for RL tasks, we hypothesize that this time-domain focus is insufficient. Instead, features in the frequency domain may reveal additional, critical insights. Robotic trajectories, for example, often exhibit properties in the frequency domain that are not easily discernible in the time domain\cite{kashiri2018overview}.

To support this hypothesis, we analyzed the training loss of Decision Diffuser\cite{ajay2022conditional} in the frequency domain using the Discrete Fourier transform (DFT). As shown in Fig.~\ref{fig1}, our analysis revealed that while high-frequency loss is relatively minor, significant and concentrated errors persist in the low-frequency range. This observation underscores a critical issue: existing time-domain-only approaches inadvertently introduce shifts in the low-frequency part of the frequency domain. Such shifts are particularly problematic because low-frequency components capture the overall trend and stability of trajectories, which are essential for smooth, regular, and optimal trajectories. Specifically, as illustrated in Fig.~\ref{epoch}, a detailed examination of the trajectory state sequence in the frequency domain reveals that its energy density distribution is predominantly concentrated in the low-frequency range. This pattern reflects the inherent continuity and smoothness in natural physical phenomena and robotic motor motion\cite{kashiri2018overview}, which stem from the physical principle of energy conservation that governs stable and efficient motion. Therefore, frequency shifts in low-frequency components may lead to outlier problems and trajectory instability.

\begin{figure*}[thpb]
  \centering
  \includegraphics[scale=0.30]{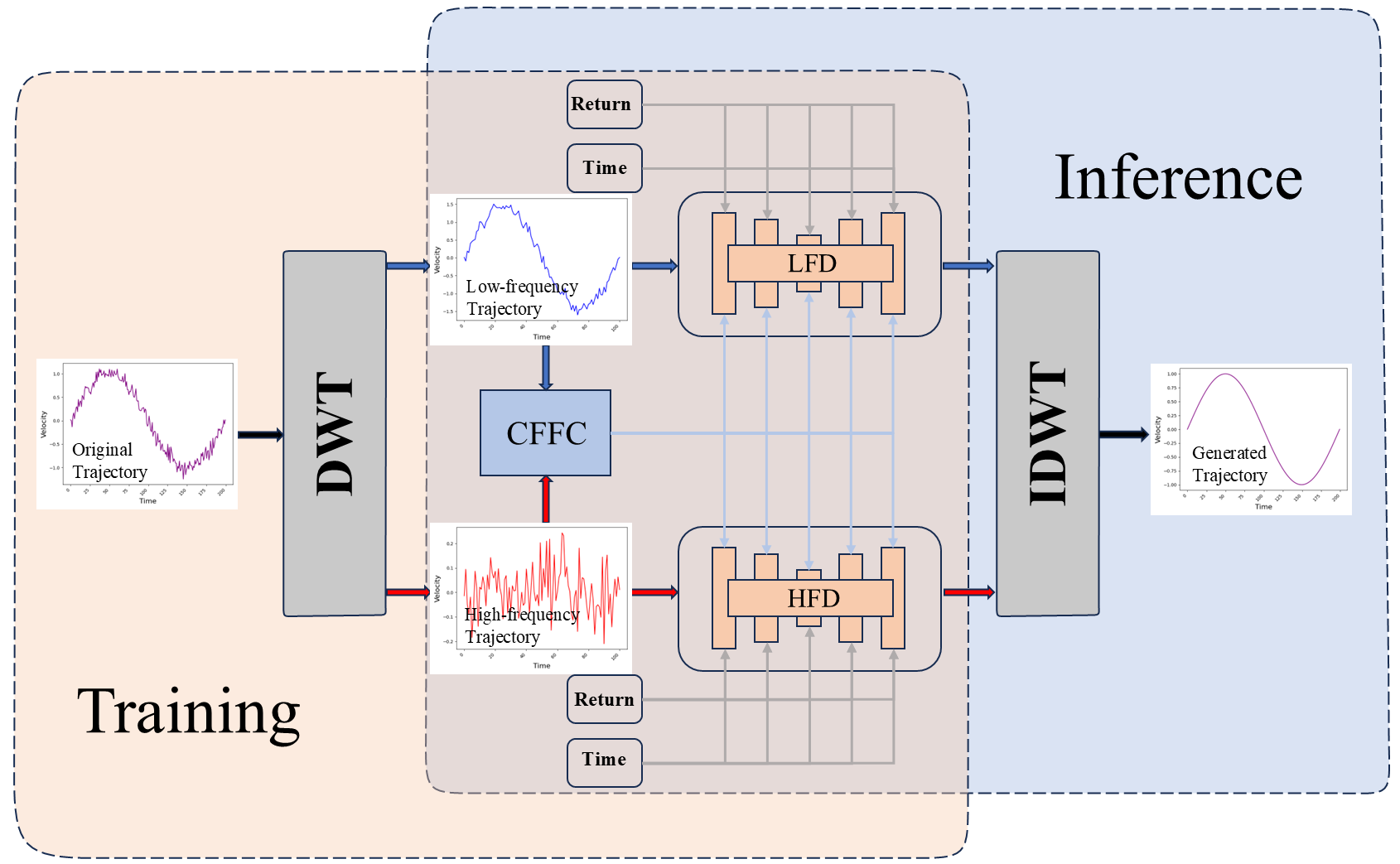}
  \caption{The overall WFDiffuser framework. During training, we apply DWT to split trajectories into low- and a high-frequency sub-trajectories. Diffusion models (LFD and HFD, low-frequency diffusion and high-frequency diffusion models) are then trained separately on each component. During inference, we utilize the IDWT to combine the generated sub-trajectories into a complete trajectory. CFFC employs STFT to extract frequency-domain features from both sub-trajectories, then applies cross attention to integrate these features as conditions for the diffusion models.}
  \label{fig2}
\end{figure*}

Building on this insight, we propose a novel diffusion-based RL framework, Wavelet Fourier Diffuser (WFDiffuser), to address the frequency shift issue, as shown in Fig.~\ref{fig2}. Our approach decomposes trajectories into low- and high-frequency components, enabling targeted diffusion modeling for each. Specifically, during training, we apply the Discrete Wavelet Transform (DWT) to split trajectories into low- and a high-frequency sub-trajectories. Low- and high-frequendcy diffusion models (LHD and HFD) are then trained separately on each component. During inference, we utilize the Inverse Discrete Wavelet Transform (IDWT) to combine the generated sub-trajectories into a complete trajectory. 

To enhance the interaction between low- and high-frequency information, WFDiffuser introduces a Cross Fourier Fusion Conditioner (CFFC) block. This block employs the Short-Time Fourier Transform (STFT) to extract frequency-domain features from both sub-trajectories, then applies cross attention to integrate these features as conditions for the diffusion models. 

We evaluate WFDiffuser on standard D4RL\cite{fu2020d4rl} tasks, and results demonstrate its effectiveness in mitigating frequency shift. Our approach achieves superior performance compared to existing approaches.  

In summary, our contributions include: 
\begin{itemize}
    \item We identify the frequency shift issue in sequence modeling-based RL and its impact on trajectory stability.
    \item We propose WFDiffuser, a diffusion-based RL framework to incorporate both wavelet and Fourier transforms for addressing frequency-domain challenges. To the best of our knowledge, WFDiffuser is the first diffusion-based RL approach that considers the frequency domain.
    \item Extensive experiments on D4RL datasets validate the efficacy of WFDiffuser, demonstrating improved performance over state-of-the-art approaches.
\end{itemize}

\section{Related Work}

\subsection{Offline RL}
Offline RL\cite{levine2020offline}\cite{prudencio2023survey} is a widely studied field where agents learns policy exclusively from offline dataset without interacting with the environment during training. 

Traditional methods\cite{kumar2020conservative}\cite{wu2019behavior}\cite{kostrikov2021offline}\cite{kostrikov2021offline1}\cite{ghosh2022offline}\cite{dadashi2021offline} address this challenge by computing policy gradients or learning a value function that estimates the discounted sum rewards from a given state. However, these approaches suffer from overestimation of the values due to the distribution differences between the offline dataset and the learned policy. To mitigate this issue, prior works have introduced action constraints\cite{yang2022behavior} and value pessimism\cite{buckman2020importance}. For example, conservative Q-learning (CQL)\cite{kumar2020conservative} enforces a conservative value function to ensure that estimated values remain lower than their true counterparts, thereby reducing overestimation bias.
\subsection{RL as Sequence Modeling}
A recent paradigm shift in RL reformulates the problem as sequence modeling, where policy are learned by directly modeling trajectory sequence autoregressively\cite{chen2021decision}\cite{janner2021offline}\cite{janner2022planning}. Notable frameworks include Decision Transformer\cite{chen2021decision} and Decision Diffuser\cite{ajay2022conditional}, which utilize transformer architectures and diffusion probability models\cite{sohl2015deep}\cite{ho2020denoising}, respectively, to capture trajectory distribution. Compared to transformer, diffusion probability models offer greater flexibility in composing constraints, making them particularly suitable for RL applications.

Based on this idea, Diffusion Q-learning (Diffusion-QL)\cite{wang2022diffusion} incorporates a regularization term in the loss function to guide the model toward learning optimal actions. Energy-guided DIffusion Sampling (EDIS)\cite{liu2024energy} leverages diffusion probability models to extract prior knowledge from the offline dataset for enhanced data generation in the online phase. Additionally, some works explore the use of diffusion probability models to improve the behavior diversity and generalization ability\cite{liang2023adaptdiffuser}\cite{liu2024didi}. For instance, AdaptDiffuser\cite{liang2023adaptdiffuser} further extends the diffusion-based approach by introducing a self-evolutionary diffusion model that generates high-quality heterogeneous data, enabling zero-shot generalization to unseen tasks.

However, existing sequence modeling-based RL methods only focus on the time-domain features, overlooking valuable frequency-domain information. We observe that this omission leads to frequency shifts, resulting in suboptimal trajectories. In contrast, our WFDiffuser integrates both time- and frequency -domain features, reducing frequency shifts and improving overall performance. 

\subsection{Diffusion Probability Models}
Diffusion probability models\cite{sohl2015deep}\cite{ho2020denoising} have demonstrated promising success in image and text generation\cite{nichol2021glide}\cite{saharia2022photorealistic} by formulating the generation process as an iterative denoising procedure. These models are closely related to energy-based models (EBMs)\cite{du2019implicit}\cite{nijkamp2019learning}, as the denoising process can be interpreted as parameterizing the gradient of the data distribution\cite{song2019generative} and optimizing the score matching object\cite{hyvarinen2005estimation}.

A growing body of efforts has explored the potential of diffusion probability models for trajectory planning\cite{janner2022planning}\cite{ajay2022conditional}\cite{wang2022diffusion}\cite{liu2024energy}\cite{liang2023adaptdiffuser}, particularly in RL. By conditioning trajectory generation on additional information\cite{ajay2022conditional}\cite{liang2023adaptdiffuser}, diffusion probability models have shown promise in producing effective behaviors, further supporting their applicability to RL.

\section{Methods}   
\subsection{Preliminaries}
RL is commonly formulated as a discounted Markov Decision Process (MDP) specified by the tuple M = ($\textit{S},\textit{A},\mathit{\mathcal{T}},\textit{R},\mathit{\rho}_0,\mathit{\gamma}$), where $\mathit{S}$ and $\mathit{A}$ denote the state space and action space respectively, $\mathit{\tau}$ represents the transition dynamics, $\textit{R}$ is the reward function, $\mathit{\rho}_0$ refers to the initial state distribution, and $\mathit{\gamma}$ denotes the discount factor. The agent interacts with the environment following a policy $\pi$, which generates a trajectory $\mathbf{\tau}$ = ($s_0,a_0,s_1,a_1,\cdots$). The objective of RL is to find a return-maximizing policy:

\begin{equation}
\pi^* = arg max_\pi E_{\mathbf{\tau}\sim\pi}[\sum_{i=0}^{\infty}\gamma^ir(s_i,a_i)]\label{eq100}
\end{equation}

As acquiring the robotics dataset in the real world can be expensive, a well-used alternative method is to utilize offline dataset D = $\{\mathbf{\tau}|\mathbf{\tau}$ $\sim$ $\pi_D (\mathbf{\tau})\}$ generated by one or more behavioral policies $\pi_D$ for offline RL. We aim to train a policy $\pi$ in \eqref{eq100} using the dataset D.

\subsection{Overall Architecture}
As illustrated in Fig.~\ref{fig2}, WFDiffuser consists of a discrete wavelet transform block, a Cross Fourier Fusion Conditioner (CFFC) block, two diffusion blocks (LFD and HFD, low- and high-frequency diffusion block), and an inverse discrete wavelet block. Inspired by previous work, we redefine the planning trajectory as a sequence of state and focus on diffusing over states rather than actions. This choice is motivated by the fact that actions exhibit higher variability, discreteness, and lower smoothness compared to states, making them more challenge to model and predict\cite{tedrake2009underactuated}. 

During training, Given a historical states sequence:
\begin{equation}
\mathbf{\tau} = (s_t, s_{t + 1}, s_{t + 2}, \cdots, s_{t + H - 1})\label{eq1}
\end{equation}

\begin{algorithm}
\caption{Inference with WFDiffuser}
\label{alg1}
\begin{algorithmic}[1]
\Require WFDiffuser (descrete wavelet transform block (DWT), inverse descrete wavelet transform block (IDWT), Cross Fourier Fusion Conditioner block (CFFC), low-frequency diffusion model (LFD), high-frequency diffusion model (HFD)), inverse dynamics $f_\phi$, history length $C$
\State Initialize $\mathbf{\tau} \leftarrow$ Queue $(length = C)$, $t \leftarrow 0$
\While {not done}
\State Observe state s; $\mathbf{\tau}$.insert(s);
\State $\mathbf{\tau}_{low}$, $\mathbf{\tau}_{high}$ = DWT ($\mathbf{\tau}$)
\State $Con_{low}$, $Con_{high}$ = CFFC ($\mathbf{\tau}_{low}$, $\mathbf{\tau}_{high}$)
\State Set $\mathbf{R(\tau)}$ = 1
\State $\mathbf{y(\tau)}_{low}$ = concatenate ($\mathbf{R(\tau)}$, $Con_{low}$) 
\State $\mathbf{y(\tau)}_{high}$ = concatenate ($\mathbf{R(\tau)}$, $Con_{high}$)
\State Sample $\mathbf{\tau}_{low}^{0}$ from LFD($\mathbf{y(\tau)}_{low}$) 
\State Sample $\mathbf{\tau}_{high}^{0}$ from LFD($\mathbf{y(\tau)}_{high}$)
\State $\mathbf{\tau}^{0}$ = IDWT ($\mathbf{\tau}_{low}^{0}$, $\mathbf{\tau}_{high}^{0}$)
\State Extract ($s_{t}$, $s_{t+1}$) from $\mathbf{\tau}^{0}$ 
\State Execute $a_t = f_\phi$$(s_t,s_{t+1})$ 
\EndWhile
\end{algorithmic}
\end{algorithm}

where t denotes the time at which a state was visited in trajectory $\mathbf{\tau}$ and we view $\mathbf{\tau}$ as a sequence of states from a trajectory of length H. The wavelet block first decomposes $\mathbf{\tau}$ into two sub-trajectories $\mathbf{\tau}_{low}$ (low-frequency components) and $\mathbf{\tau}_{high}$ (high-frequency components) based on DWT. The CFFC block then facilitates cross-frequency interaction between these two sub-trajectories. Consequently, $\mathbf{\tau}_{low}$ and $\mathbf{\tau}_{high}$ are processed by two separate diffusion blocks, using the cross-frequency interaction from CSFC block as a condition. The goal is to model the data distribution under the cross-frequency condition. Since actions are not included in the diffusion process, we train an inverse dynamics model to predict the action between two adjacent states. Notably, the inverse dynamics model is trained on the original trajectory $\mathbf{\tau}$, rather than its wavelet-transformed components.

During inference, we first sample $\mathbf{\tau}_{low}^{0}$ and $\mathbf{\tau}_{high}^{0}$ from the respective diffusion blocks, conditioned on the CFFC output based on the historical states. The inverse wavelet block then reconstructs the complete trajectory $\mathbf{\tau}^{0}$ via IDWT. Finaly, actions are predicted and executed autoregressively between adjacent states in $\mathbf{\tau}^{0}$. The full inference procedure is outlined in Algorithm \ref{alg1}.

\subsection{Discrete Wavelet Transform}
DWT has been widely applied in low-level tasks, as it enables transformation from the time domain to the wavelet domain, where key properties become more apparent. Here we adopt Haar Wavelets as the mother wavelet to decompose the trajectory. Given a trajectory $\mathbf{\tau}$ define in\eqref{eq1}, we apply DWT to obtain:
\begin{equation}
\mathbf{\tau}_{low},\mathbf{\tau}_{high} = DWT(\mathbf{\tau})\label{eq2}
\end{equation}

\begin{figure}[htbp]
\centerline{\includegraphics[scale=0.2]{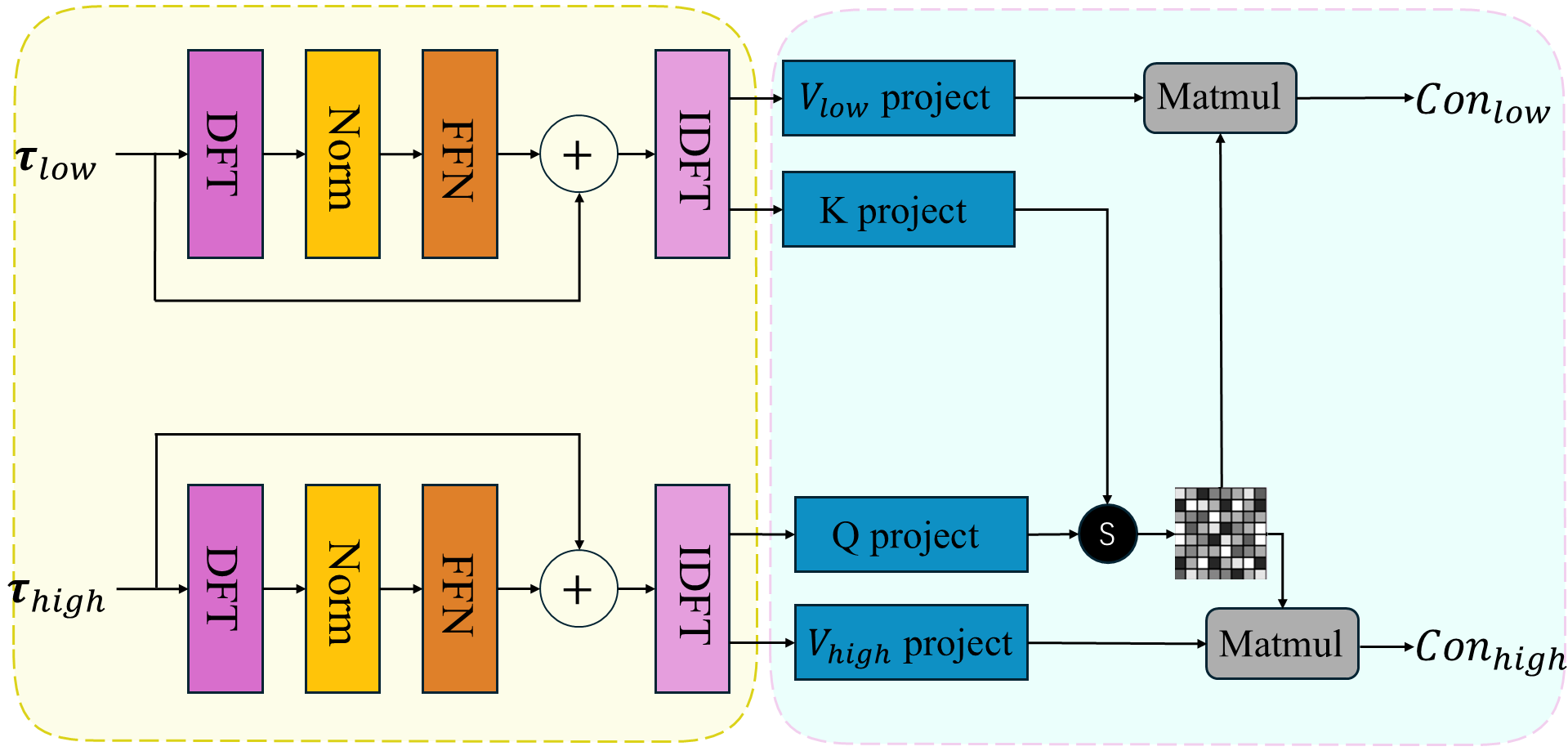}}
\caption{The structure of Cross Fourier Fusion Conditioner (CFFC) block. It consists of a fourier frequency module (the left yellow part) and a cross attention module (the right blue part). S refers to the softmax operation. }
\label{cffc}
\end{figure}

Haar wavelets consist of a low-pass filter L and a high-pass filter H, as follows:
\begin{equation}
\begin{aligned}
L = \frac{1}{\sqrt{2}}[1,1]^T,
H = \frac{1}{\sqrt{2}}[1,-1]^T\label{eq3}
\end{aligned}
\end{equation}

We first perform a down-sampling convolution on $\mathbf{\tau}$ by computing the average of the sum of each pair of adjacent states. The result is then passed through the low-pass filter in \eqref{eq3} to get $\mathbf{\tau}_{low}$:
\begin{equation}
\mathbf{\tau}_{low} = (s_t^{low}, s_{t + 1}^{low}, s_{t + 2}^{low}, \cdots, s_{t + \frac{H}{2} - 1}^{low})\label{eq4}
\end{equation}

Similarly, we perform another down-sampling convolution on $\mathbf{\tau}$, but this time, we compute the average of the difference between each pair of adjacent states. The result is the passed through the high-pass filter in\eqref{eq3} to get $\mathbf{\tau}_{high}$:
\begin{equation}
\mathbf{\tau}_{high} = (s_t^{high}, s_{t + 1}^{high}, s_{t + 2}^{high}, \cdots, s_{t + \frac{H}{2} - 1}^{high})\label{eq5}
\end{equation}

Here, $\mathbf{\tau}_{low}$ captures the global trend of the trajectory $\mathbf{\tau}$, while $\mathbf{\tau}_{high}$ represents local variations. Due to the biorthogonal property of DWT, the sub-trajectories retain all information despite down-sampling.

\subsection{Cross Fourier Fusion Conditioner}
Before applying diffusion to $\mathbf{\tau}_{low}$ and $\mathbf{\tau}_{high}$, we introduce the Cross Fourier Fusion Conditioner (CFFC) block to enable information exchange between these sub-trajectories. In this subsection, we delve into the mechanism of CFFC, which is trainable and leverages Short-Time Fourier Transform (STFT) and cross attention to efficiently fuse features between sub-trajectories, across time and frequency domain. The detailed structure of CFFC is shown in Fig.~\ref{cffc}. To be specific, CFFC is composed of a fourier frequency module and a cross attention module. 

\begin{figure}[b]
\centerline{\includegraphics[scale=0.35]{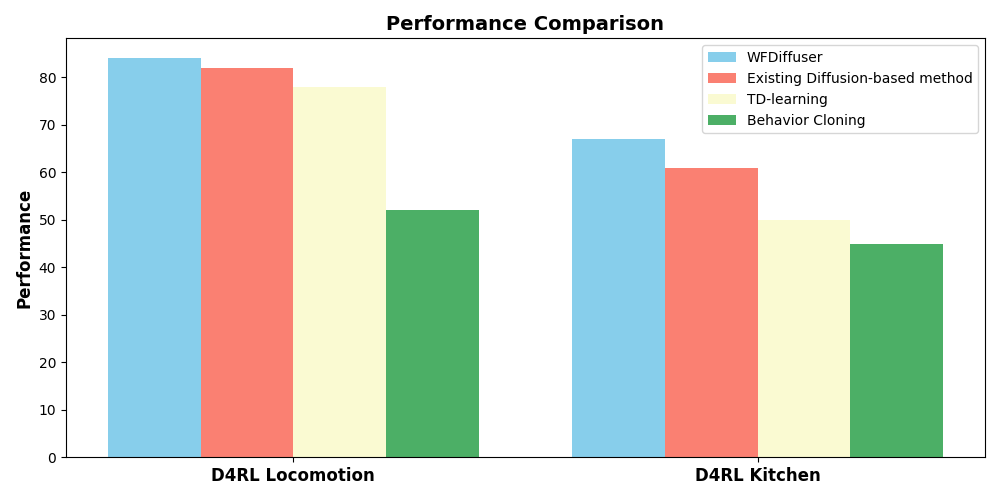}}
\caption{ WFDiffuser performs better than existing diffuser-based (Decision Diffuser\cite{ajay2022conditional}), TD-learning (CQL\cite{kumar2020conservative}), and Behavior Cloning (BC\cite{torabi2018behavioral}). The performance score refers to the normalized average returns on D4RL benchmark\cite{fu2020d4rl}.}
\label{exp1}
\end{figure}

The fourier frequency module aims to enhance features by incorporating frequency-domain information. Given an input sequence $\mathbf{\tau}_{low}$ in \eqref{eq4}, the computation process is as follows (a similar process is applied to $\mathbf{\tau}_{high}$):
\begin{equation}
\begin{aligned}
A_{low}, P_{low} = DFT((\mathbf{\tau}_{low}))\\
A^{'}_{low} = FFN(Norm(A_{low})) + A_{low}\\
P^{'}_{low} = FFN(Norm(P_{low})) + P_{low}\\
\mathbf{\tau}^{'}_{low} = IDFT(A^{'}_{low}, P^{'}_{low})\label{eq6}
\end{aligned}
\end{equation}
Where FFN is a feed-forward network and Norm refers to normalization. The DFT and IDFT denote the Discrete Fourier Transform and its inverse (for simplicity, here we let $t = 0$ and denote $n = \frac{H}{2}$ in \eqref{eq4}):
\begin{equation}
\begin{aligned}
\textit{F}(\mathbf{\tau}_{low})(k) = \sum_{i=0}^{n - 1} s^{low}_i exp(-\frac{j2\pi}{n}ki), 0\leq k \leq n\\
\textit{F}^{-1}(\mathbf{F}_{low}(i) = \frac{1}{n}\sum_{k=0}^{n - 1} F^{low}_k exp(\frac{j2\pi}{n}ki), 0\leq i \leq n\label{eq7}
\end{aligned}
\end{equation}
where $j = \sqrt{-1}$ is the imaginary unit. The A and P represent the amplitude and phase components in the frequency domain, respectively.   

While the fourier frequency module fuses the features across time and frequency domain for
$\mathbf{\tau}_{low}$ and $\mathbf{\tau}_{high}$ respectively, the following cross attention module aims at facilitating mutual reinforcement between these sub-trajectories. Here we use linear projections of $\mathbf{\tau}^{'}_{high}$ to construct Q and $\mathbf{\tau}^{'}_{low}$ to construct K:
\begin{equation}
\begin{aligned}
Q = FFN(\mathbf{\tau}^{'}_{high})\\
K = FFN(\mathbf{\tau}^{'}_{low})\label{eq8}
\end{aligned}
\end{equation}

Similarly, $V_{low}$ and $V_{high}$ can be obtained by:
\begin{equation}
\begin{aligned}
V_{low} = FFN(\mathbf{\tau}^{'}_{low})\\
V_{high} = FFN(\mathbf{\tau}^{'}_{high})\label{eq9}
\end{aligned}
\end{equation}

The output feature as conditions to diffusion blocks can then be obtained from the formula:
\begin{equation}
\begin{aligned}
Con_{low} = Softmax(\frac{QK^{T}}{\sqrt{d_k}})V_{low}\\
Con_{high} = Softmax(\frac{QK^{T}}{\sqrt{d_k}})V_{high}\label{eq10}
\end{aligned}
\end{equation}
where $\sqrt{d_k}$ refers to the number of columns of Q.

\subsection{Conditional Diffusion}

The diffusion process is applied separately to $\mathbf{\tau}_{low}$ and $\mathbf{\tau}_{high}$, respectively. For simplicity, we only demonstrate the diffusion process of $\mathbf{\tau}_{low}$ here, while a similar process is applied to $\mathbf{\tau}_{high}$.   

The purpose of the diffusion process is to predict future states that maximize the reward-to-go. Following previous work, we treat this as a conditional diffusion problem, where the diffusion is applied over states and is formulated as:
\begin{equation}
q(\mathbf{\tau}_{low}^{i+1}|\mathbf{\tau}_{low}^{i})\quad\quad p_\theta(\mathbf{\tau}_{low}^{i-1}|\mathbf{\tau}_{low}^{i},\mathbf{y(\tau)}_{low})\label{eq11}
\end{equation}

Here, q represents the predefined forward noising process, which adds Gauss noise to the original trajectory, while p denotes the trainable reverse denoising process. The index i indicates the diffusion timestep, where $\mathbf{\tau}_{low}^{0}$ refers to the original low-frequency sub-trajectory in the forward noising process and a generated sample in the reverse denoising process.

The new term $\mathbf{y(\tau)}_{low}$ in \eqref{eq11} serves as the conditional information given to the reverse denoising process. Specifically, we define it as:
\begin{equation}
\mathbf{y(\tau)}_{low} = concatenate(Con_{low},\mathbf{R(\tau)})\label{eq12}
\end{equation}
where $\mathbf{R(\tau)}$  represents the normalized returns of the original trajectory $\mathbf{\tau}$, and $Con_{low}$ is the output conditional feature from CFFC. The returns are normalized to keep $\mathbf{R(\tau)} \in [0,1]$. The inclusion of $\mathbf{y(\tau)}_{low}$ in the reverse denoising process ensures that the generated trajectories exhibit properties aligned with the given conditional information. This design allows us to generate cross-compatible $\mathbf{\tau}_{low}$ and $\mathbf{\tau}_{high}$, maximizing returns while ensuring that their composition results in an optimal trajectory. Without this conditioning, it is possible that both low- and high-frequency sub-trajectories achieve high returns in conditions but fail to form an optimal trajectory when combined.

Specifically, we employ classifier-free guidance with low-temperature sampling in the reverse diffusion process in \eqref{eq11}. The update step for $\mathbf{\tau}_{low}$ is formulated as:
\begin{equation}
\mathbf{\tau}_{low}^{i-1} = \mathbf{\tau}_{low}^{i} - \hat{\epsilon}\label{eq13}
\end{equation}
where the perturbed noise $\hat{\epsilon}$ is computed as:
\begin{equation}
\hat{\epsilon} = \epsilon_{\theta}(\mathbf{\tau}_{low}^i,\emptyset,i) + \omega(\epsilon_{\theta}(\mathbf{\tau}_{low}^i,\mathbf{y(\tau)}_{low},i) - \epsilon_{\theta}(\mathbf{\tau}_{low}^i,\emptyset,I))\label{eq14}
\end{equation}
Here, The scalar $\omega$ is designed to amplify the most favorable trajectories that align with $\mathbf{y(\tau)}_{low}$.

During inference, we compute $Con_{low}$ and $Con_{high}$ based on historical states, then sample $\mathbf{\tau}_{low}^{0}$ and $\mathbf{\tau}_{high}^{0}$ from the reverse denoising process with $\mathbf{R(\tau)}$ = 1. Consequently, we reconstruct the full trajectory using IDWT:
\begin{equation}
\mathbf{\tau}^0 = IDWT(\mathbf{\tau}_{low}^{0}, \mathbf{\tau}_{high}^{0})\label{eq15}
\end{equation}

Finally, actions are generated based on the adjacent states in $\mathbf{\tau}^0$ from the learned inverse dynamics model. We are then able to generate a policy.

\section{Experiments}

\setlength{\tabcolsep}{3pt}  

\begin{table*}[htbp]\tiny
    \centering
    \caption{Offline Reinforcement Learning performance.}
    \resizebox{\textwidth}{!}{
    \small
    \begin{tabular}{ll*{8}{c}}
        \toprule
        \textbf{Dataset} & \textbf{Environment} & \textbf{BC} & \textbf{CQL} & \textbf{IQL} & \textbf{DT} & \textbf{TT} & \textbf{MOReL} & \textbf{Decision Diffuser} & \textbf{WFDiffuser} \\
        \midrule
        \multirow{3}{*}{Med-Expert} 
        & HalfCheetah & 55.2 & 91.6 & 86.7 & 86.8 & \textbf{95} & 53.3 & 90.6 & 92.8 $\pm$ 1.1 \\
        & Hopper & 52.5 & 105.4 & 91.5 & 107.6 & 110.0 & 108.7 & 111.8 & \textbf{114.7} $\pm$ 1.3 \\
        & Walker2d & 107.5 & 108.8 & \textbf{109.6} & 101.9 & 95.6 & 108.4 & 108.8 & \textbf{108.6} $\pm$ 1.8 \\
        \midrule
        \multirow{3}{*}{Medium} 
        & HalfCheetah & 42.6 & 44.0 & 47.4 & 42.6 & 46.9 & 42.1 & 49.1 & \textbf{51.5} $\pm$ 0.9 \\
        & Hopper & 52.9 & 58.5 & 66.3 & 67.6 & 61.1 & \textbf{95.4} & 79.3 & 84.5 $\pm$ 2.8 \\
        & Walker2d & 75.3 & 72.5 & 78.3 & 74.0 & 79.0 & 77.8 & 82.5 & \textbf{86.1} $\pm$ 1.2 \\
        \midrule
        \multirow{3}{*}{Med-Replay} 
        & HalfCheetah & 36.6 & \textbf{45.5} & 44.2 & 36.6 & 41.9 & 40.2 & 39.3 & 38.1 $\pm$ 2.2 \\
        & Hopper & 18.1 & 95.0 & 94.7 & 82.7 & 91.5 & 93.6 & 100 & \textbf{103.4} $\pm$ 2.6 \\
        & Walker2d & 26.0 & 77.2 & 73.9 & 66.6 & \textbf{82.6} & 49.8 & 75.0 & 76.3 $\pm$ 3 \\
        \midrule
        \textbf{Average} &  & 51.9 & 77.6 & 77.0 & 74.7 & 78.9 & 72.9 & 81.8 & \textbf{84} \\
        \midrule
        \multirow{2}{*}{Mixed} 
        & Kitchen & 51.5 & 52.4 & 51.0 & - & - & - & 65 & \textbf{69.4} $\pm$ 3.2 \\
        & Partial Kitchen & 38.0 & 50.1 & 46.3 & - & - & - & 57 & \textbf{64.1} $\pm$ 2.6 \\
        \midrule
        \textbf{Average} &  & 44.8 & 51.2 & 48.7 & - & - & - & 61 & \textbf{66.8} \\
        \bottomrule    
    \end{tabular}    
    }   
    \vspace{1mm} 

    { 
    $^{\mathrm{a}}$ We report the mean and the standard error over 5 random seeds.\\
    $^{\mathrm{b}}$ The performance score refers to the normalized average returns on D4RL tasks\cite{fu2020d4rl}.
    }
     
    \label{tab1}
\end{table*}

In this section, we present the experimental setup and performance evaluation of WFDiffuser across various RL task (illustrated in Fig.~\ref{exp1}). Beyond assessing our method's ability to generate effective RL policies, we also investigate its capability to mitigate frequency shift in the frequency domain. Additionally, we empirically validate the use of CFFC and compare different mother wavelets.

\subsection{Experiment setup}
We evaluate WFDiffuser in four distinct environments: HalfCheetah, Hopper, Walker2D and Kitchen. The datasets we used are sourced from the publicly available D4RL benchmark\cite{fu2020d4rl}. For the three locomotion environments (HalfCheetah, Hopper, Walker2D), we use three types of dataset: medium, medium-replay, and medium-expert. For the kitchen manipulation environment, we experiment with both tasks-mixed and tasks-partial datasets.

We conduct a comparative analysis against a broad range of offline RL baselines, including model-free algorithms such as CQL\cite{kumar2020conservative}, model-based algorithms such as Model-Based Offline Reinforcement Learning (MOReL)\cite{kidambi2020morel}, and sequence models like Decison Transformer\cite{chen2021decision} and Decison Diffuser\cite{ajay2022conditional}. 

In our implementation, we represent the noise model $\epsilon_{\theta}$  in the reverse denoising process using a temporal U-net, constructed following previous work, and we train $\epsilon_{\theta}$ using the Adam optimizer\cite{kingma2014adam} with a learning rate of $2 \times 10^{-4}$. Additionly, the inverse dynamics and the FFNs in the SFFC block are all implemented as a 2-layered MLP with 512 hidden units and ReLU activations. We use the horizon H of 96.

\begin{figure}[htbp]
\centerline{\includegraphics[scale=0.5]{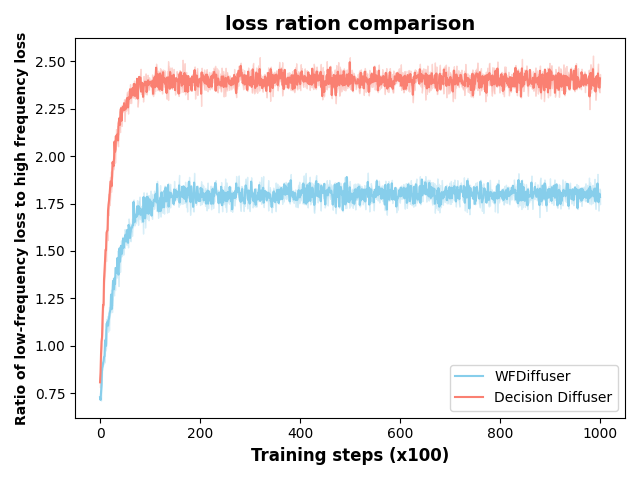}}
\caption{WFDiffuser successfully mitigates the frequency shift in the low-frequency range during training, which results in more stable trajectories and better performance. The dataset we use is Hopper-medium-v2}
\label{exp2}
\end{figure}

\subsection{Main Results}
The quantitative results of our experiments are presented in table.~\ref{tab1}. Our findings show that WFDiffuser is either competitive with or outperforms many offline RL baselines. Notably, the performance improvement between WFDiffuser and other methods is most significant in the D4RL Kitchen tasks, which are particularly challenging due to their requirement for long-term decision making. 

\begin{figure}[htbp]
\centerline{\includegraphics[scale=0.4]{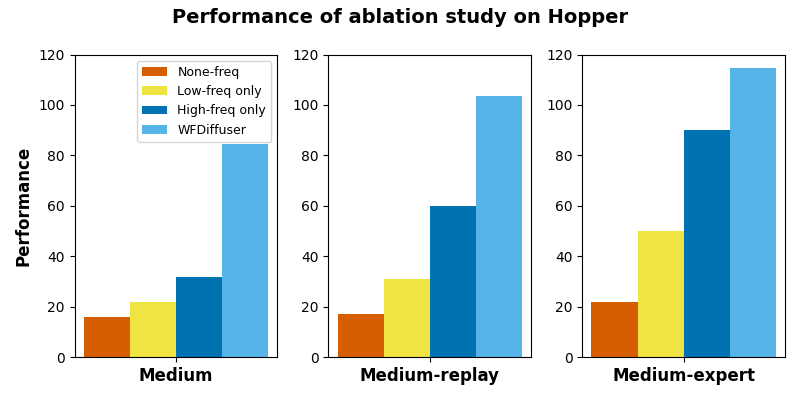}}
\caption{Ablation study on the Hopper. The performance score refers to the normalized average returns on D4RL benchmark\cite{fu2020d4rl}.}
\label{exp3}
\end{figure}

\setlength{\tabcolsep}{3pt}  

\begin{table}[htbp]
    \centering
    \caption{The impact of mother wavelet.}
    \resizebox{\columnwidth}{!}{
    \begin{tabular}{ll*{8}{c}}
        \toprule
        \textbf{Environment} & \textbf{Dataset} & \textbf{Daubechies} & \textbf{Morlet} & \textbf{Haar} \\
        \midrule
        \multirow{3}{*}{Hopper} 
        & Med-Expert & 114.5 $\pm$ 5.2 & 114.2 $\pm$ 3.7 & \textbf{114.7} $\pm$ 1.3\\
        & Medium & 84.1 $\pm$ 3.1 & 83.5 $\pm$ 3.0 & \textbf{84.5} $\pm$ 2.8\\
        & Med-Replay & 103.4 $\pm$ 5.7 & 103.1 $\pm$ 2.4 & \textbf{103.4} $\pm$ 2.6\\
        \midrule
        \textbf{Average} &  & 100.7 & 100.4 & \textbf{100.9} \\ 
        \bottomrule
    \end{tabular}    
    }  
    \vspace{1mm} 

    { 
    $^{\mathrm{a}}$ We report the mean and the standard error over 5 random seeds.\\
    $^{\mathrm{b}}$ The performance score refers to the normalized average returns on D4RL tasks\cite{fu2020d4rl}.
    }
    \label{tab3}
\end{table}

To further investigate whether WFDiffuser successfully mitigates the frequency shift problem discussed in section~\ref{intro}, we analyze the loss during training and transform it into the frequency domain using DWT. We define the first and last 10 frequency modes as the low-frequency and high-frequency components, respectively. We then compute the loss ratio between these two parts and compare WFDiffuser against Decision Diffuser. As demonstrated in Fig.~\ref{exp2}, WFDiffuser effectively reduces frequency shifts in the low-frequency part, which ensures smoother and more stable trajectories, ultimately leading to higher returns.

\subsection{Ablation Study}
To assess the impact of the CFFC block in WFDiffuser, we conduct an ablation study on the environment of Hopper. Specifically, we define 
three variants:

\begin{itemize}
    \item \textbf{Low-freq-only} Applies conditional features from CFFC only to low-frequency diffusion block. Note that despite the condition information being passed to the low-frequency diffusion block, the useful condition information actually originates from the high-frequency component.
    \item \textbf{High-freq-only}
    Applies conditional features from CFFC only to high-frequency diffusion block. Note that despite the condition information being passed to the high-frequency diffusion block, the useful condition information actually originates from the low-frequency component.
    \item \textbf{None-freq}
    Does not apply frequency-based conditional features to either diffusion block.
\end{itemize}

The result illustrated in Fig.~\ref{exp3} shows that WFDiffuser outperforms all three variants in all cases, highlighting the importance of frequency interaction between low-frequency and high-frequency components. Additionally, we observe that high-freq-only consistently outperforms low-freq-only, demonstrating the crucial role of low-frequency information. While a sufficient low-frequency sub-trajectory can be generated without frequency interaction, the high-frequency sub-trajectory requires guidance from low-frequency information to prevent mismatched sub-trajectory generation.

Furthermore, we investigate the impact of different mother wavelets in the discrete wavelet transform block. The mother wavelet is responsible for decomposing trajectory sequences into low- and high-frequency components. We select the Haar Wavelet as our default choice due to its simplicity and ease of deployment. However, it is important to note that Haar Wavelet is not the only viable option. Alternative wavelets, such as the Daubechies Wavelet or Morlet Wavelet, can also be used for decomposition. To assess their effectiveness, We conduct an additional ablation study, replacing Haar wavelet with these wavelets. As shown in table.~\ref{tab3}, while these alternatives perform reasonably well, Haar wavelet consistently yields the best result. Therefore, we adopt the Haar wavelet as default. 

\section{Conclusion}

In this paper, we propose WFDiffuser, an innovative diffusion-based RL framework that emphasizes frequency domain analysis. WFDiffuser introduces a novel frequency-based structure, which effectively decomposes trajectory sequences into distinct frequency components and diffuse models them with cross-frequency interaction. By addressing the frequency shift in the frequency domain, our method ensures the generation of smooth and stable trajectories achieving higher returns. This design empowers WFDiffuser to enhance decison-making through frequency domain interpolation. Our results provide a new perspective on viewing RL from a frequency domain analysis standpoint and underscore its potential.

For future work, applying WFDiffuser to real-world tasks presents an exciting direction. Compared to simulation tasks, real-world scenarios often involve more high-frequency noise, which makes WFDiffuser theoretically more advantageous as it explicitly incorporates frequency information into decision-making.

\section*{Acknowledgment}

This work was supported by the Natural Science Foundation of Shenzhen (No. JCYJ20230807111604008, No. JCYJ20240813112007010), the Natural Science Foundation of Guangdong Province (No.2024A1515010003), National Key Research and Development Program of China (No. 2022YFB4701400) and Cross-disciplinary Fund for Research and Innovation (No. JC2024002) of Tsinghua SIGS.

\bibliographystyle{IEEEtran}
\bibliography{main}

\vspace{12pt}

\end{document}